\documentclass[letterpaper,10pt,conference]{ieeeconf}
\IEEEoverridecommandlockouts
\usepackage{amsmath,amssymb}
\usepackage{graphicx}
\usepackage{booktabs}
\usepackage{array}
\usepackage{cite}
\usepackage{microtype}
\usepackage{xcolor}
\usepackage{svg}
\usepackage{placeins}
\usepackage{float}
\usepackage{needspace}
\usepackage[hidelinks]{hyperref}
\newcommand{\methodname}{Uni-VLaT}
\newcommand{\draftfig}[2]{%
  \fbox{\begin{minipage}[c][#1][c]{0.96\linewidth}
  \centering\small\textbf{#2}
  \end{minipage}}%
}

\title{\methodname{}: Whole-Body Tactile Adaptation of VLA Policies for Humanoid Loco-Manipulation}

\author{
Zihao Wang$^{1,*}$,
Shutong Liu$^{2,*}$,
Siqi Zheng$^{3}$,
Liu Cao$^{1}$,
Ruoqu Chen$^{1}$,
Rundong Liu$^{1}$,
Yanchao Yang$^{4}$,
Mengdi Xu$^{1,\dagger}$%
\thanks{
$^{1}$Tsinghua University;
$^{2}$Beihang University;
$^{3}$Communication University of China;
$^{4}$The University of Hong Kong;
$^{*}$Equal contribution.
$^{\dagger}$Corresponding author.
}%
}

\begin{document}
\bstctlcite{BSTcontrol}
\IEEEaftertitletext{%
  \vspace{-1.2em}
  \begin{minipage}{\textwidth}
    \includegraphics[width=\textwidth,trim=0 0 0 8bp,clip]{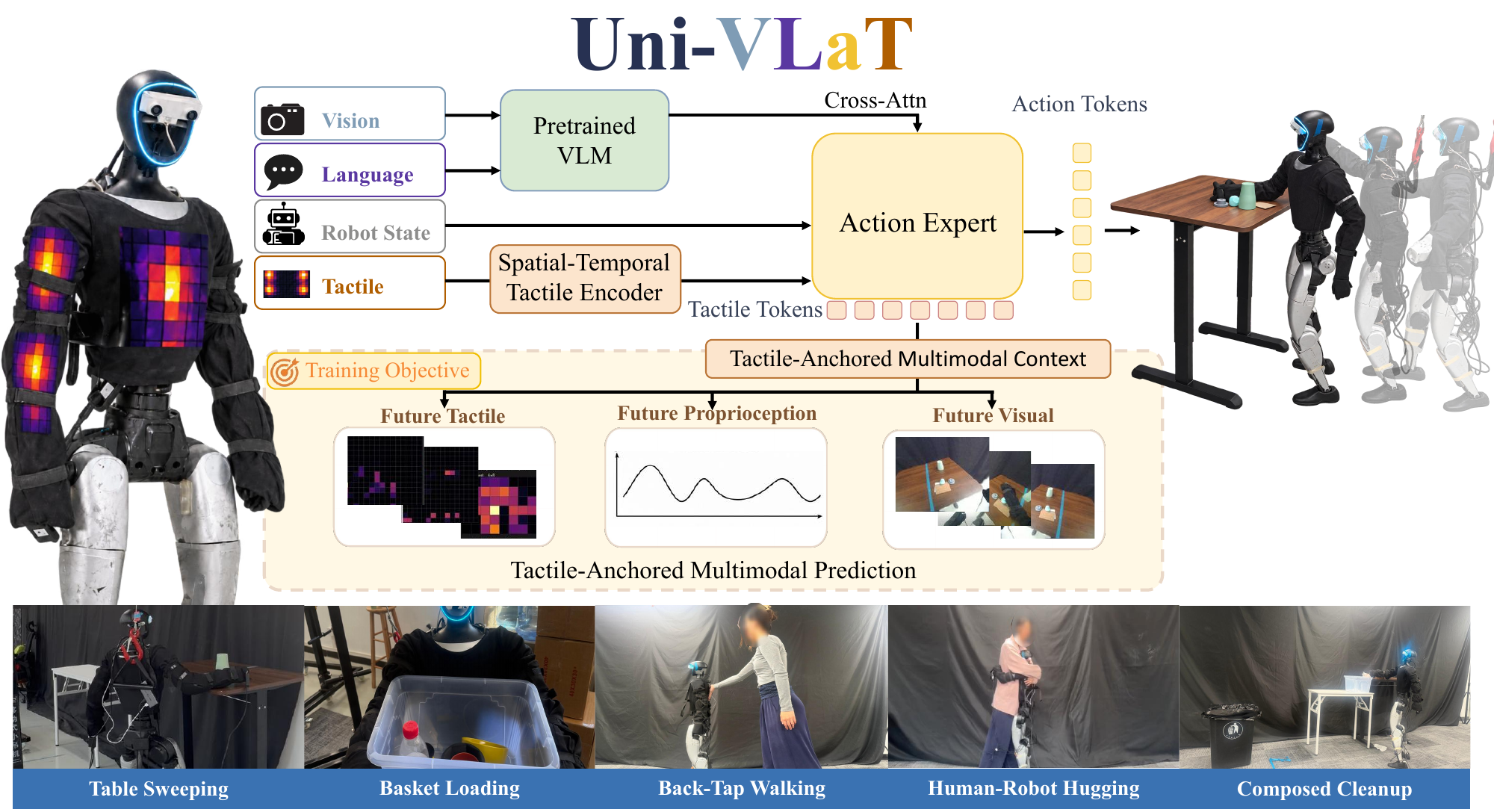}
    \refstepcounter{figure}\label{fig:teaser}
    \vspace{0.2em}\par
    {\small\textbf{Fig.~\thefigure.} \methodname{} adapts pretrained VLA policies with distributed whole-body tactile sensing and tactile-anchored future representation prediction for contact-rich humanoid loco-manipulation across five real-world tasks.}
  \end{minipage}
  \vspace{0.5em}
}
\maketitle

\begin{abstract}
Physical contact often determines how a humanoid should respond during loco-manipulation, yet vision and proprioception alone are often insufficient to characterize physical interaction, especially when the contact region is occluded. Unlike sparse force or torque measurements at predefined regions, distributed tactile sensing preserves spatially resolved contact patterns across the robot body. We therefore study how to integrate such whole-body tactile information into vision-language-action (VLA) policies for contact-rich control. Our approach, \methodname{}, introduces a tactile pathway whose latent state is trained not only for action generation, but also to predict future tactile, proprioceptive, and visual representations. This predictive objective builds a tactile-anchored multimodal context, encouraging a more structured understanding of the physical world. We evaluate \methodname{} on five real-robot tasks covering tactile-triggered locomotion, sustained physical interaction, human--robot contact, and loco-manipulation. \methodname{} achieves a 75\% average success rate, outperforming a baseline without tactile input by 43 points and a tactile-input baseline without predictive supervision by 7 points. Across two pretrained VLA backbones, our method improves Table Sweeping by 30 points on both backbones and Back-Tap Walking by 85--90 points. Ablations further show that contextualized tactile prediction and absolute future targets are critical to performance. These results indicate that predictive tactile learning provides an effective route for extending pretrained VLA policies to whole-body physical interaction. Project page: \href{https://ggkiller-air.github.io/Uni-VLaT/}{\textcolor{blue}{Uni-VLaT.github.io/}}
\end{abstract}

\section{Introduction}
\label{sec:introduction}

Pretrained vision-language-action (VLA) models increasingly provide reusable visual, linguistic, and action priors across robot tasks and embodiments~\cite{rt2,openx,octo,openvla,pi0,pi05,grootn1}. This capability is particularly valuable for humanoids, whose large workspaces and whole-body action spaces make task-specific policy learning expensive. Recent systems have extended pretrained or generalist policies to humanoid loco-manipulation~\cite{humanoidvla,sonic,wholebodyvla,openhlm}, yet these policies primarily perceive interaction through visual and proprioceptive observations~\cite{beyondsight}. In many learning-based robot control systems, policies generate motion or position references that are tracked by low-level controllers. Without direct tactile or force observations, physical contact is regulated only implicitly: small pose errors or changing loads can produce insufficient or excessive contact, making sustained interaction difficult to stabilize and potentially unsafe~\cite{tact,wtumi,htd}. Distributed whole-body tactile sensing closes this feedback gap by directly revealing where contact occurs and how it evolves across body regions, including broad or occluded interactions that vision cannot reliably resolve.

Effective tactile adaptation, however, requires more than appending another sensor stream. Whole-body tactile signals are spatially distributed, intermittent, and action-dependent; their control meaning varies with task intent, body state, and ongoing motion. Moreover, pretrained VLAs were not trained to align such signals with their semantic and action representations. Simply concatenating tactile features therefore does not ensure that the policy will relate contact to the corresponding body response and scene evolution. The central challenge is how to enable pretrained VLA policies to effectively use whole-body tactile feedback for contact-rich control.

Recent work has explored distributed tactile sensing for humanoid manipulation. TACT incorporates visual and tactile observations into whole-body imitation learning, WT-UMI uses tactile/force feedback for contact-aware planning and control, and HTD trains a dedicated multimodal Transformer with future tactile-latent and force prediction~\cite{tact,wtumi,htd}. These works establish the value of whole-body tactile, but leave two aspects underexplored in our setting. First, they largely rely on dedicated policy or planning architectures, leaving unclear how distributed tactile can augment pretrained generalist robot policies while building on their existing semantic and action priors. Second, when predictive supervision is used, it is primarily centered on future tactile or force signals rather than jointly characterizing how contact evolves together with body state and visual observations.

Our key idea is to use tactile as a \emph{physical anchor} for multimodal interaction modeling. Tactile directly identifies physical interaction at the robot body, while proprioception captures the associated body response and vision captures changes in the surrounding scene. We therefore construct a tactile-anchored multimodal context from contextualized tactile features and train it to predict future tactile, proprioceptive, and visual representations in their respective latent spaces. Without naively fusing the modality-specific representations into a single undifferentiated representation, this formulation uses tactile to connect their complementary observations of the same physical interaction.

We instantiate this idea in \textbf{\methodname{}}, a framework for adapting pretrained humanoid VLA policies with tactile sensing. Rather than treating tactile as an additional input modality alone, \methodname{} uses tactile features as a physical anchor for multimodal interaction modeling. After interacting with visual, language, proprioceptive, and action representations inside the pretrained policy, the tactile features form a tactile-anchored multimodal context. From this context, three modality-specific predictors estimate future tactile, proprioceptive, and visual representations. By jointly predicting how contact, body state, and scene appearance evolve, \methodname{} encourages the policy to learn a representation of physical interaction that is directly relevant to control.

Our contributions are three-fold:
\begin{itemize}

\item \textbf{Tactile-anchored multimodal learning.}
We introduce \methodname{}, which learns a tactile-anchored multimodal context through auxiliary predictive supervision. This context is used to predict future tactile, proprioceptive, and visual representations, encouraging the policy to capture the coupled evolution of physical interaction.

\item \textbf{Tactile Adaptation of Pretrained VLAs.}
We introduce a tactile adaptation strategy that leverages pretrained VLA semantic, visual, and action priors to ground tactile in task context and support contact-rich control without retraining the policy from scratch.

\item \textbf{Extensive humanoid evaluation.}
We evaluate \methodname{} on five Unitree G1 tasks against multiple baselines, across two pretrained VLA backbones and a from-scratch policy comparison, with controlled ablations of the tactile and predictive components.

\end{itemize}

\begin{figure*}[!t]
    \centering
    \includegraphics[width=\textwidth]{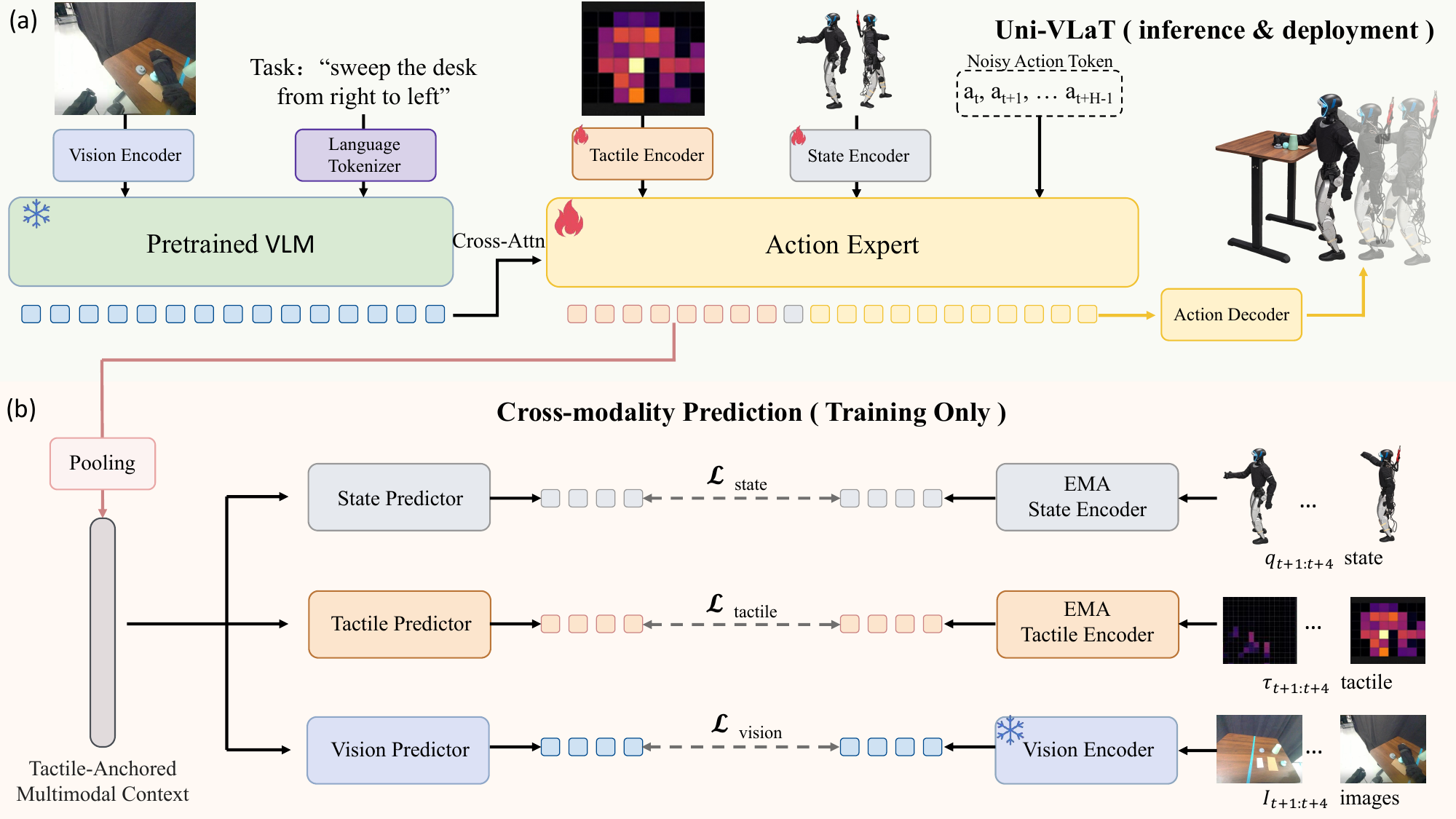}
    \caption{Overview of \methodname{}. Whole-body tactile features are integrated into a pretrained VLA policy for action generation, while training-time multimodal prediction is supervised by future tactile, proprioceptive, and visual representations.}
    \label{fig:pipeline}
\end{figure*}

\section{Related Work}
\label{sec:related}

\paragraph{Pretrained robot policies}
Pretrained robot policies provide reusable visual, linguistic, and action priors. Gato, RT-1, PaLM-E, and RoboCat established generalist control and embodied grounding~\cite{gato,rt1,palme,robocat}. Open X-Embodiment supplies cross-robot data, whereas RT-2 transfers web knowledge to actions~\cite{openx,rt2}. Octo and OpenVLA provide generalist formulations~\cite{octo,openvla}; RDT-1B, $\pi_0$, and $\pi_{0.5}$ scale generative control~\cite{rdt1b,pi0,pi05}; and GR00T N1 targets humanoids~\cite{grootn1}. Diffusion Policy, ACT, Mobile ALOHA, and FAST develop denoising, action chunks, whole-body data, and efficient tokenization~\cite{diffusionpolicy,act,mobilealoha,fast}. CogACT, X-VLA, and ReconVLA refine policy representations~\cite{cogact,xvla,reconvla}, while StarVLA and DiT4DiT address modularity and video--action modeling~\cite{starvla,dit4dit}. Yet distributed contact remains weakly represented.

\paragraph{Tactile representation and VLA adaptation}
Tactile policy learning requires effective sensing and integration with pretrained policies. DIGIT provides high-resolution sensing~\cite{digit}; Touch and Go pairs touch with vision~\cite{touchandgo}; and Sparsh learns self-supervised tactile representations~\cite{sparsh}. TACTO supports simulation, while ManiFeel benchmarks visuotactile policies~\cite{tacto,manifeel}. Beyond Sight performs heterogeneous-sensor adaptation through language grounding~\cite{beyondsight}. Tactile-VLA connects touch to physical knowledge through VLA finetuning~\cite{tactilevla}. VLA-Touch uses dual-level feedback for planning and action refinement~\cite{vlatouch}. TacCoRL introduces sim--real co-training~\cite{taccorl}, whereas HapticVLA distills tactile-aware behavior without inference-time touch~\cite{hapticvla}. These works mainly study localized contact.

\paragraph{Humanoid whole-body interaction}
HumanoidBench, OmniH2O, HumanPlus, and HOVER advance whole-body learning~\cite{humanoidbench,omnih2o,humanplus,hover}. SONIC learns latent control~\cite{sonic}; Humanoid-VLA and LeVERB add visual or language conditioning~\cite{humanoidvla,leverb}; and WholeBodyVLA and OpenHLM address loco-manipulation~\cite{wholebodyvla,openhlm}. TACT adds upper-body touch to imitation learning~\cite{tact}. WT-UMI uses force-supervised planning~\cite{wtumi}, while HTD predicts tactile latents and hand forces~\cite{htd}. \methodname{} instead adapts pretrained policies for distributed touch.

\paragraph{Predictive representations for physical interaction}
Predictive representations model dynamics without raw reconstruction. I-JEPA and V-JEPA predict image and video embeddings~\cite{ijepa,vjepa}; V-JEPA 2 and DINO-WM extend prediction to planning~\cite{vjepa2,dinowm}. DreamVLA and Cosmos Policy predict future knowledge or visual dynamics~\cite{dreamvla,cosmospolicy}, while VLA-JEPA and JEPA-WAM target action-relevant states~\cite{vlajepa,jepawam}. Dream-Tac models action and visual--tactile futures~\cite{dreamtac}. Tactile-WAM uses asymmetric attention~\cite{tactilewam}; VT-WAM predicts visual, tactile, and action information~\cite{vtwam}; and TacWAM adds mechanics-aware prediction~\cite{tacwam}. HTD predicts tactile latents and hand forces~\cite{htd}. \methodname{} anchors multimodal future supervision in contextualized tactile states.

\section{Problem Formulation}
\label{sec:problem}

We study language-conditioned humanoid loco-manipulation with distributed whole-body tactile feedback. At observation step $t$, the policy receives
\begin{equation}
    \mathcal{O}_t=(I_t,\ell,q_t,\tau_{t-H_T+1:t}), 
    \tau_{t-H_T+1:t}=(\tau_{t-3},\ldots,\tau_t),
    \label{eq:observation}
\end{equation}
where $I_t$ denotes egocentric stereo RGB images, $\ell$ a language instruction, $q_t$ the proprioceptive state available to the policy, and $\tau_t$ the distributed tactile measurements. We use a causal history of $H_T=4$ tactile frames.

Our action space is the 64-dimensional latent motion space of SONIC~\cite{sonic}. A pretrained VLA policy is adapted to generate a chunk of motion tokens through flow matching, which are passed to SONIC's whole-body control decoder:
\begin{equation}
\begin{aligned}
    A_t &= (a_{t},\ldots,a_{t+H_A-1})
    \sim \pi_\theta(\cdot\mid\mathcal{O}_t), \\
    q^{\mathrm{des}} &= \mathcal{D}_{\mathrm{SONIC}}(s,a),
    \qquad a\in A_t.
\end{aligned}
\label{eq:policy_to_control}
\end{equation}
Here $A_t\in\mathbb{R}^{H_A\times64}$ is a chunk of 64-dimensional latent motion tokens with \(H_A=40\) in all experiments. At each control update, $a$ denotes the token selected for execution from the active chunk, and $s$ denotes SONIC's controller observation, including joint positions and velocities, base inertial measurements, and previous control outputs. The pretrained control decoder $\mathcal{D}_{\mathrm{SONIC}}$ maps the selected motion token and robot feedback to desired joint positions $q^{\mathrm{des}}$, which are tracked by joint-level PD controllers. We directly supply the predicted tokens to this decoder through SONIC's Protocol v4 interface, enabling coordinated upper- and lower-body execution through a single whole-body controller.

Given synchronized demonstrations $\mathcal{D}=\{(\mathcal{O}_t,A_t^\star)\}$, where $A_t^\star$ is the demonstrated motion-token chunk, our objective is to learn a tactile-conditioned VLA policy that generates whole-body motion from visual, linguistic, proprioceptive, and tactile observations. While adapting the VLA policy, the pretrained SONIC controller serves as the low-level execution interface.

\section{Method}
\label{sec:method}

\subsection{Method Overview}
\methodname{} adapts a pretrained VLA policy through tactile integration and multimodal predictive supervision (Fig.~\ref{fig:pipeline}). A spatial encoder and a temporal module transform distributed tactile measurements into tokens summarizing recent contact. These tokens interact with vision-language, proprioceptive, and action features within the policy's DiT trunk. The action-token outputs generate motion commands, while the contextualized tactile-token outputs form a tactile-anchored multimodal context for predicting future tactile, proprioceptive, and visual representations. We keep the pretrained VLM frozen and fine-tune the action expert, state encoder, and tactile modules. Action supervision and the three predictive objectives jointly train the tactile-conditioned policy. The predictive heads and EMA target encoders are used only during training.

\subsection{Tactile Encoding and Policy Integration}

\textbf{Spatial tactile encoding.}
Our tactile sensors cover eight body regions: the chest, central back, left and right shoulders, left and right upper back, and left and right arms. We suppress sensor noise with a fixed deadband and normalize readings by region. For the tactile sleeves, we pool measurements around the arm circumference to reduce sensitivity to sleeve rotation while preserving contact location along the arm.

Let $\tau_{t,r}$ denote the tactile measurements for region $r$ at time $t$ after the preprocessing described above. Each region has an independent MLP $E_r$ that maps its measurements to a common feature dimension, producing the regional embeddings $R_t=[E_r(\tau_{t,r})]_{r=1}^{8}$. To combine contact information across regions, we introduce a set of eight learned queries $Q^T$ and use the regional embeddings as both keys and values:
\begin{equation}
    S_t^T=\operatorname{LN}\!\left(
    \operatorname{CrossAttn}(Q^T,R_t,R_t)\right),
    \label{eq:tactile_spatial}
\end{equation}
where the three attention arguments denote queries, keys, and values, respectively, and $\operatorname{LN}$ denotes layer normalization. The superscript $T$ denotes the tactile modality. The output $S_t^T$ contains eight tactile tokens, each of which can aggregate information from multiple body regions. The learned queries are shared across frames, providing a consistent token ordering for temporal encoding.

\textbf{Temporal contact encoding.}
A single tactile observation may be insufficient to distinguish contact establishment, sustained support, and release. We therefore use a lightweight temporal Transformer to aggregate the spatial tactile tokens over a four-frame history:
\begin{equation}
    Z_t^T=\Phi_T(S_{t-3}^T,S_{t-2}^T,S_{t-1}^T,S_t^T).
    \label{eq:tactile_temporal}
\end{equation}
Here $\Phi_T$ applies temporal attention to the four-frame sequence at each tactile-token index, using learned time embeddings to encode frame order. The aggregated features are combined with the current-frame tokens through a residual connection, producing temporally informed tactile tokens $Z_t^T$. A learned scalar gate scales these tokens before they enter the pretrained policy.

\textbf{Integration into the action head.}
The VLA action head uses a DiT trunk to jointly process the gated tactile tokens, one proprioceptive token, and noisy action tokens. Visual and language features from the pretrained VLM are supplied through cross-attention. Self-attention enables interaction among tactile, proprioceptive, and action tokens, while cross-attention incorporates scene and instruction information. This contextualizes contact with the robot's body state and task, while making tactile information available for action generation.

The action-token outputs are projected to flow-matching velocity predictions, which are used to generate the motion-token chunk through iterative integration. The tactile-token outputs, denoted by $h^T_{t,1},\ldots,h^T_{t,8}$, are retained for the predictive supervision described next. Both outputs come from the same DiT trunk, linking tactile contextualization to the action-generation pathway.

\begin{figure*}[!t]
    \centering
    \includegraphics[width=\textwidth]{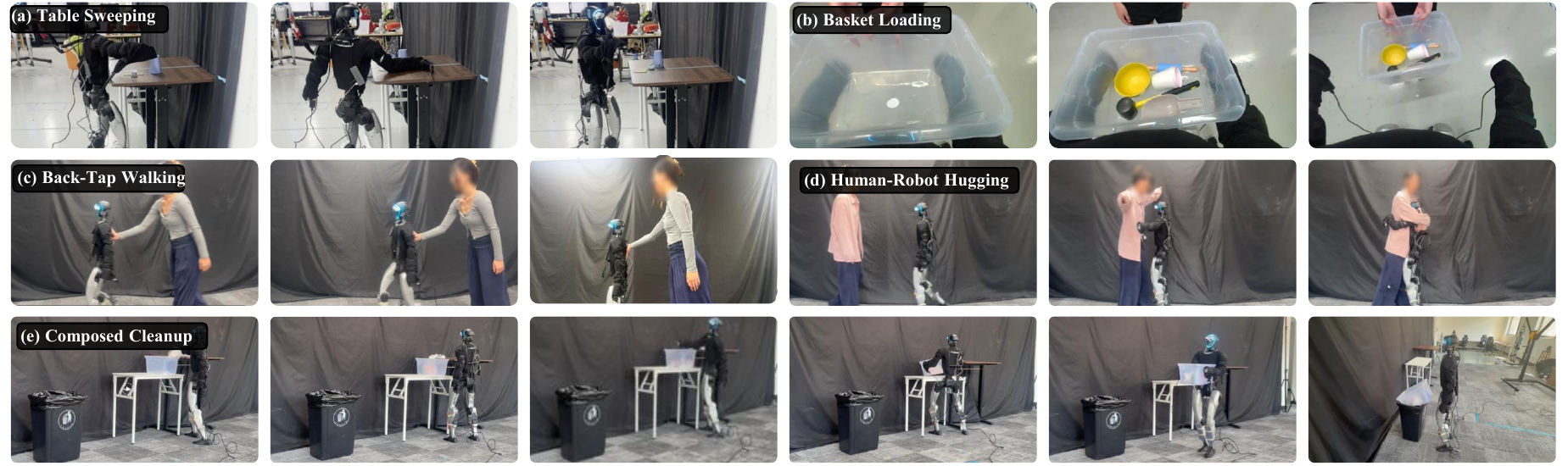}
    \caption{Real-robot evaluation tasks: (a) Table Sweeping, (b) Basket Loading, (c) Back-Tap Walking, (d) Human--Robot Hugging, and (e) Composed Cleanup.}
    \label{fig:real_robot_tasks}
\end{figure*}

\subsection{Tactile-Anchored Multimodal Prediction}

\textbf{Predictive context.}
Through self-attention and cross-attention in the DiT trunk, tactile tokens incorporate information from proprioceptive, action, and vision-language features. The resulting post-DiT tactile representations therefore provide a tactile-anchored context for predicting how the interaction evolves across modalities. To obtain a compact input for the predictive heads, we average these contextualized tokens:
\begin{equation}
    c_t=\frac{1}{8}\sum_{n=1}^{8}h^T_{t,n}.
    \label{eq:contact_context}
\end{equation}
Here $c_t$ is a compact readout of the \emph{tactile-anchored multimodal context} learned through policy attention. Mean pooling reduces the predictor input size. Three modality-specific MLP predictors map $c_t$ to future latent sequences:
\begin{equation}
    \widehat z^m_{t+1:t+K}=P_m(c_t),\qquad
    K=4,\quad m\in\{T,P,V\},
    \label{eq:predictors}
\end{equation}
where $T$, $P$, and $V$ denote tactile, proprioceptive, and visual modalities. Each predictor jointly outputs a sequence of four latents, one for each future observation step.

\textbf{Future target representations.}
Let \(E_T\), \(E_P\), and \(E_V\) denote the spatial tactile, proprioceptive, and visual encoders, respectively. We use EMA target encoders \(\bar E_T\) and \(\bar E_P\), while \(E_V\) remains frozen. The prediction targets are encoded from the corresponding future observations:
\begin{align}
    z^T_{t+k}&=\operatorname{sg}\!\left(\bar{E}_T(\tau_{t+k})\right),\nonumber\\
    z^P_{t+k}&=\operatorname{sg}\!\left(\bar{E}_P(q_{t+k})\right),\nonumber\\
    z^V_{t+k}&=\operatorname{sg}\!\left(E_V(I_{t+k})\right),
    \label{eq:absolute_target}
\end{align}
where \(k=1,\ldots,K\) and \(\operatorname{sg}\) denotes stop-gradient.
The tactile target encoder processes each future frame independently and mean-pools its spatial tokens, while the proprioceptive target encoder embeds the future robot state. For vision, the frozen pretrained encoder extracts patch features that are averaged within each image and across camera views. Each modality provides one target latent per future step.

All branches predict absolute future latents in their respective modality-specific spaces. These targets retain both persistent interaction states, such as sustained contact during box support, and changes in contact, posture, and scene appearance. Joint prediction encourages \(c_t\) to capture these complementary aspects of the interaction without naively merging the three modalities into a single representation.

\textbf{Joint objectives and gradient flow.}
Following HTD~\cite{htd}, we combine cosine-based directional alignment with SmoothL1 regression of latent magnitudes, applying this objective to each of the three predicted modalities:
\begin{align}
    \mathcal{L}_m=\frac{1}{K}\sum_{k=1}^{K}\Big[&1-\cos(\widehat z^m_{t+k},z^m_{t+k})\nonumber\\
    &+\beta\operatorname{SmoothL1}\!\left(
    \|\widehat z^m_{t+k}\|_2,\|z^m_{t+k}\|_2\right)\Big],
    \label{eq:latent_loss}
\end{align}
where $\beta=1$. The cosine term aligns feature directions, while the SmoothL1 term constrains feature magnitudes. The overall training objective is
\begin{equation}
    \mathcal{L}=\mathcal{L}_{\mathrm{act}}
    +\lambda_T\mathcal{L}_T+\lambda_P\mathcal{L}_P+\lambda_V\mathcal{L}_V,
    \label{eq:total_loss}
\end{equation}
where $\mathcal{L}_{\mathrm{act}}$ is the flow-matching velocity MSE over valid action dimensions. We use $\lambda_T=0.05$ and $\lambda_P=\lambda_V=0.005$ across tasks.

Action-loss gradients propagate through the action-token outputs and the shared DiT attention pathways to the trainable tactile encoding modules. Auxiliary-loss gradients pass through the three predictors and $c_t$ to the post-DiT tactile tokens, updating the predictors and the trainable policy and tactile modules along these paths. Thus, tactile tokens receive supervision both from demonstrated actions and from future multimodal observations. Stop-gradient prevents predictive losses from updating the target encoders directly; tactile and proprioceptive teachers instead follow their online encoders through EMA updates, while the visual teacher remains frozen.

\textbf{Deployment.}
At deployment, a rolling buffer supplies the four-frame tactile history, which is encoded into tokens for the tactile-conditioned VLA policy. The policy generates chunks of 64-dimensional motion tokens, and the token selected at each control update is sent to SONIC's control decoder through its Protocol v4 interface. As defined in Eq.~\ref{eq:policy_to_control}, the decoder combines this token with robot proprioceptive feedback to produce joint-position targets, which are tracked by joint-level PD controllers for coordinated upper- and lower-body execution. The predictive heads and EMA target encoders are omitted at deployment.

\begin{table*}[!t]
\centering
\caption{Real-robot success rates (\%) with Isaac-GR00T.}
\label{tab:main_success_rate}
\small
\renewcommand{\arraystretch}{1.15}
\begin{tabular*}{\textwidth}{@{\extracolsep{\fill}}lcccc@{}}
\toprule
\textbf{Task}
& \textbf{No Tactile}
& \textbf{Tactile w/o Pred.}
& \textbf{Tactile Prediction}
& \textbf{\methodname{} (Ours)} \\
\midrule
Back-Tap Walking      & 0\%  & 85\%          & \textbf{90\%} & 85\%          \\
Table Sweeping        & 45\% & 60\%          & 60\%          & \textbf{75\%} \\
Basket Loading        & 30\% & 65\%          & 70\%          & \textbf{80\%} \\
Human--Robot Hugging  & 55\% & \textbf{80\%} & 75\%          & \textbf{80\%} \\
Composed Cleanup      & 30\% & 50\%          & 50\%          & \textbf{55\%} \\
\midrule
\textbf{Avg.}         & 32\% & 68\%          & 69\%          & \textbf{75\%} \\
\textbf{Avg. w/o Back-Tap} & 40.0\% & 63.8\% & 63.8\% & \textbf{72.5\%} \\
\bottomrule
\end{tabular*}
\end{table*}

\begin{figure*}[!t]
    \centering
    \IfFileExists{figures/Fig4_CrossBackbone.pdf}{
        \includegraphics[width=1\textwidth]{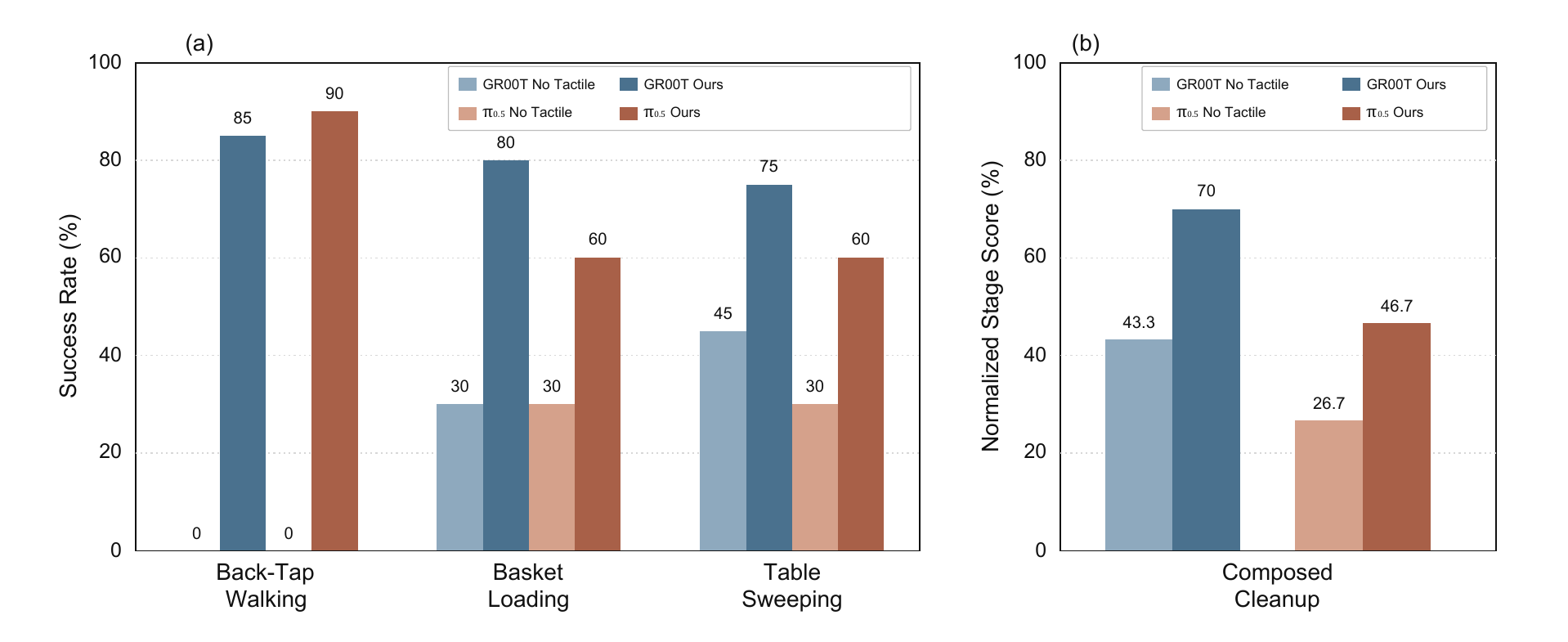}
    }{
        \draftfig{1.75in}{Cross-policy evaluation across four tasks}%
    }
    \caption{Cross-policy evaluation: success rates (left) and the three-stage Composed Cleanup score (right).}
    \label{fig:backbone_comparison}
\end{figure*}

\section{Experiments}
\label{sec:experiments}

We validate \methodname{} by answering three questions:

\textbf{Q1:} Does whole-body tactile sensing improve contact-rich control, and can multimodal prediction further improve over tactile input w/o prediction?

\textbf{Q2:} Does tactile adaptation transfer across pretrained VLA backbones, and how does it compare with training from scratch?

\textbf{Q3:} How do the predictive context and future-target formulation affect performance?

\subsection{Experimental Setup}
\label{sec:exp_setup}

\textbf{Robot platform.}
We conduct all real-robot experiments on a Unitree G1 humanoid equipped with a stereo RGB camera and distributed tactile sensors over the torso, shoulders, upper back, and arms. The tactile system uses custom textile electronic skin from JQ Industries, based on piezoresistive fiber pressure sensors that provide spatially resolved contact measurements. Whole-body motion is executed through the pretrained SONIC controller~\cite{sonic}. The policy receives egocentric stereo RGB observations, a language instruction, proprioceptive state, and distributed tactile measurements.

\textbf{Demonstrations and training.}
We collect approximately 50 demonstrations per task. Within each comparison, methods share demonstrations, action labels, and the downstream interface. Detailed dataset statistics are provided in Appendix~\ref{app:data_collection}. Isaac-GR00T, $\pi_{0.5}$, and the ablations train for 50,000 steps; DP trains to convergence.

\textbf{Evaluation protocol.}
We report success rate (SR) over real-robot rollouts. Main-table configurations use 20 rollouts. Cross-policy Isaac-GR00T success rates reuse that evaluation; each $\pi_{0.5}$ success-rate configuration uses 10 rollouts. Ablations reuse the 20-rollout Full \methodname{} results, while other variants use 10. Instability, falling, an emergency stop, or any human intervention counts as failure. DP was trained on the same demonstrations and action interface; three checkpoints from different training stages all failed SONIC's pre-execution safety check, so its SR is reported as N/A. For cross-policy evaluation on Composed Cleanup, we additionally report a normalized stage score to capture partial progress: each rollout is scored on three milestones---sweeping rubbish into the box, lifting the box, and carrying it to the waste bin---each worth one third of the score. 

\textbf{Compared methods.}
For the main comparison, we evaluate four variants that progressively introduce tactile sensing and predictive supervision.
\emph{No Tactile} uses vision, language, and proprioception without tactile input or auxiliary predictive objectives.
\emph{Tactile w/o Pred.} adds the same tactile observation pathway as \methodname{} but is trained only with the action objective.
\emph{Tactile Prediction} additionally predicts future tactile representations while setting $\lambda_P=\lambda_V=0$.
\textbf{\methodname{}} jointly predicts future tactile, proprioceptive, and visual representations.
All four variants retain the same action-learning objective.

\subsection{Experimental Tasks}
\label{sec:exp_tasks}

We evaluate five contact-rich humanoid behaviors covering tactile-triggered locomotion, contact-guided manipulation, changing physical load, human--robot interaction, and multi-stage loco-manipulation.

\textbf{Back-Tap Walking.}
A person contacts the robot's back to trigger walking. Success requires forward motion within 1\,s after contact and stopping within 1\,s after release.

\textbf{Table Sweeping.}
The robot sweeps five objects across a tabletop center line under varying table heights. All five must cross within 60\,s; dropping any object is failure.

\textbf{Basket Loading.}
The robot supports a basket while a person sequentially places objects inside and then removes the basket. Success requires maintaining stable support without dropping the basket or its contents, and releasing appropriately when the person takes the basket away.

\textbf{Human--Robot Hugging.}
The robot embraces people of different sizes. It must complete the encircling motion, release both arms after the person disengages, and avoid unsafe contact or intervention.

\textbf{Composed Cleanup.}
The robot gathers all tabletop rubbish into a container, carries it to a waste bin, and deposits it with its contents. Any failed stage makes the rollout unsuccessful.

\subsection{Main Results}
\label{sec:main_results}

\textbf{Overall task performance.}
Table~\ref{tab:main_success_rate} reports the success rates of the four methods across all five tasks. Removing tactile sensing results in an average success rate of only 32\%, showing the limitation of vision and proprioception alone for sustained or partially occluded physical interaction. Introducing tactile input substantially improves the average success rate to 68\%, confirming that direct observation of physical contact is important for these behaviors.

Future tactile prediction provides a further but relatively modest improvement, reaching 69\% average success rate. In comparison, \methodname{} achieves the best performance of 75\%. When Back-Tap Walking is excluded, the corresponding average SRs are 40.0\%, 63.8\%, 63.8\%, and 72.5\%, with \methodname{} retaining an 8.7-point gain over Tactile w/o Pred. The gain over tactile input w/o prediction indicates that the benefit of our method does not arise solely from providing additional tactile information to the policy. Instead, jointly predicting future tactile, proprioceptive, and visual representations provides additional supervision for learning how contact, body response, and scene evolution are coupled during interaction.

The largest gains of \methodname{} over tactile input are observed on Table Sweeping and Basket Loading, where successful behavior depends on adapting to evolving physical interaction rather than only detecting whether contact has occurred. On Table Sweeping, \methodname{} improves success from 60\% to 75\%, while on Basket Loading it improves from 65\% to 80\%. Back-Tap Walking represents a tactile-as-instruction setting: No Tactile achieves 0\% SR, whereas all tactile-enabled variants reach 85--90\%. This suggests that direct tactile sensing is sufficient when contact primarily conveys a discrete task instruction, while predictive multimodal supervision is more beneficial when contact must be interpreted together with evolving body state and scene dynamics.

\textbf{Cross-policy evaluation.}
Figure~\ref{fig:backbone_comparison} extends cross-policy evaluation to four tasks: Table Sweeping, Back-Tap Walking, Basket Loading, and Composed Cleanup. On Isaac-GR00T, \methodname{} improves success rate from 45\% to 75\% on Table Sweeping and from 30\% to 80\% on Basket Loading, while maintaining the strong gain on Back-Tap Walking (0\% to 85\%). On $\pi_{0.5}$, \methodname{} similarly improves Table Sweeping from 30\% to 60\%, Basket Loading from 30\% to 60\%, and Back-Tap Walking from 0\% to 90\%.

For Composed Cleanup, binary end-to-end success is too coarse for comparing partial progress on this long-horizon task, especially when some backbone-policy combinations rarely complete the full sequence. We therefore report the normalized stage score defined above. Under this metric, \methodname{} improves Isaac-GR00T from 43.3 to 70.0 and $\pi_{0.5}$ from 26.7 to 46.7. These results show that tactile adaptation transfers across pretrained VLA backbones not only for tactile-triggered behaviors, but also for sustained contact-rich manipulation and multi-stage loco-manipulation. The from-scratch DP baseline remains N/A because its outputs failed SONIC's pre-execution safety check.

\textbf{Contact interaction analysis.}
Success rate alone does not reveal how tactile-conditioned policies respond during continuous physical interaction. We therefore further analyze Basket Loading, where the contact state changes repeatedly as objects are added to the basket. Tactile measurements are recorded for analysis in all rollouts but are withheld from the No Tactile policy input.

We summarize the right-arm tactile response as
\begin{equation}
y(t)=\frac{1}{256}\sum_{c=1}^{256}\max(x_c(t)-2,0),
\end{equation}
where $x_c(t)$ is the baseline-subtracted ADC reading of taxel $c$, and 2 is an additional deadband for suppressing small sensor noise. Thus, \(y(t)\) is an average tactile response in arbitrary ADC units; it reflects changes in contact response but does not directly measure force.

\begin{figure}[!htbp]
    \centering
    \includegraphics[width=\columnwidth]{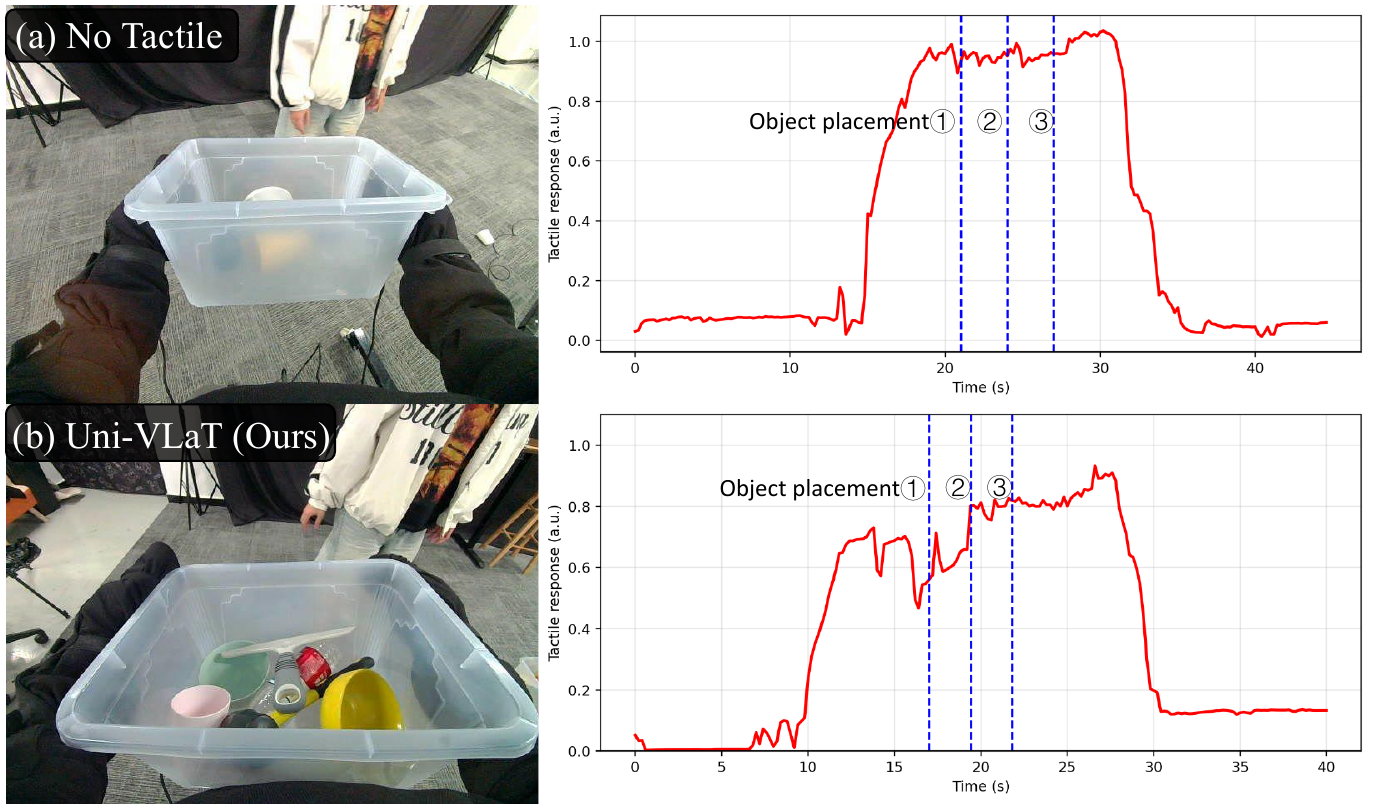}
    \caption{Basket Loading contact dynamics. Tactile response is aligned with object placements and final removal. Values are arbitrary units.}
    \label{fig:basket_dynamics}
\end{figure}

As shown in Fig.~\ref{fig:basket_dynamics}, No Tactile establishes a relatively large contact response from the outset and shows little systematic change as objects are added. In contrast, \methodname{} maintains a gentler contact level and responds to successive loading events as the basket becomes heavier. This suggests that \methodname{} uses tactile feedback to adapt its interaction to changing physical load, rather than maintaining a fixed contact pattern.

\subsection{Ablation Study}
\label{sec:ablation}

We conduct controlled ablations on Table Sweeping and Back-Tap Walking using Isaac-GR00T to isolate the effects of the predictive context and future-target formulation.

\emph{Multimodal-Input Prediction} uses visual, tactile, and proprioceptive features to predict future visual, tactile, and proprioceptive representations.
\emph{Pre-DiT Tactile Prediction} instead uses only the pre-DiT tactile representations as the source for all three predictors.
These two variants examine whether prediction from the contextualized post-DiT tactile states is beneficial.
Finally, \emph{Delta Prediction Targets} replaces the absolute future latent targets with delta targets, $\Delta z_{t+k}^{m}=z_{t+k}^{m}-z_t^{m}$, while retaining the same three prediction modalities and loss formulation.
All variants use the same four-frame tactile history and retain the action-learning objective.

\begin{table}[!t]
\centering
\caption{Ablation results with Isaac-GR00T.}
\label{tab:ablation}
\small
\renewcommand{\arraystretch}{1.25}

\begin{tabular*}{\columnwidth}{@{\extracolsep{\fill}}lccc@{}}
\toprule
\textbf{Variant}
& \textbf{Sweeping}
& \textbf{Back-Tap}
& \textbf{Avg.} \\
\midrule
\textbf{Full \methodname{}}
& 75\% & 85\% & 80\% \\
Multimodal-Input Prediction
& 40\% & 80\% & 60\% \\
Pre-DiT Tactile Prediction
& 30\% & 80\% & 55\% \\
Delta Prediction Targets
& 20\% & 50\% & 35\% \\
\bottomrule
\end{tabular*}
\end{table}

Full \methodname{} achieves 80\% average SR, compared with 60\% for Multimodal-Input Prediction and 55\% for Pre-DiT Tactile Prediction. The gap is most pronounced on Table Sweeping, where the three variants achieve 75\%, 40\%, and 30\%, respectively, while Back-Tap Walking remains similar at 85\%, 80\%, and 80\%. These results first show that incorporating visual and proprioceptive information is important for interpreting tactile signals beyond contact alone. Moreover, the stronger performance of the post-DiT tactile context suggests that grounding prediction in a contextualized tactile representation better captures multimodal interaction, whereas direct multimodal prediction may rely on modality-specific shortcuts.

Delta targets reduce average SR from 80\% to 35\%. Future--current differences can be small in latent space and easily obscured by representation noise or prediction error, while absolute targets remain directly anchored to the encoder space and preserve persistent contact and body-state information.

\section{Conclusion and Limitations}
\label{sec:conclusion}

We presented \methodname{}, a tactile adaptation framework that uses tactile as a physical anchor to connect visual, proprioceptive, and tactile information through future representation prediction. By building on pretrained VLA semantic, visual, and action priors, \methodname{} grounds whole-body tactile feedback in task context without learning the policy from scratch. Across five real-robot tasks, \methodname{} achieves 75\% average SR, outperforming No Tactile by 43 points and Tactile w/o Pred. by 7 points. Cross-policy evaluation and controlled ablations further support the benefits of pretrained priors, post-DiT tactile-anchored context, and absolute future targets.

Our experiments also expose several limitations of current whole-body tactile learning. The array sensors remain sensitive to noise and mounting variation, motivating more transferable sensing and representations beyond fixed deadbands, normalization, and axial pooling. Limited whole-body tactile simulation further restricts large-scale robustness studies. Finally, our evaluation covers both tactile-as-instruction behaviors, such as Back-Tap Walking, and tasks requiring sustained contact adaptation. The former mainly benefits from the availability of tactile input, whereas the latter places greater demands on how tactile feedback is interpreted and integrated for control. Due to the constraints of our laboratory setup, we evaluate this distinction with five representative tasks. Future work should extend this setting to richer whole-body interaction tasks that require finer tactile regulation and more precise contact control.

\FloatBarrier

\section*{Acknowledgment}

This work is supported in part by the Dushi Program. The views and opinions expressed in this work are solely those of the authors.

\bibliographystyle{IEEEtran}
\bibliography{refs}
\clearpage

\appendices
\renewcommand{\thesection}{\Alph{section}.}
\renewcommand{\thesectiondis}{\Alph{section}.}
\makeatletter
\renewcommand{\section}{\@startsection{section}{1}{\z@}%
  {1.5ex plus 1.5ex minus 0.5ex}{0.7ex plus 1ex minus 0ex}%
  {\normalfont\normalsize\bfseries\raggedright}}
\makeatother

\begin{center}
\bfseries\large APPENDIX
\end{center}
\vspace{0.2em}

\section{Data Collection and Dataset}
\label{app:data}

\subsection{Demonstration Collection}
\label{app:data_collection}

We collect real-robot demonstrations on a Unitree G1 using PICO-based VR teleoperation together with the pretrained SONIC whole-body controller. The operator's teleoperation commands are converted by SONIC into a 64-dimensional latent motion token, which is then decoded by the whole-body controller into joint-position targets. During data collection, we record the 64-dimensional \texttt{action.motion\_token} as the high-level action label, together with stereo RGB observations, proprioceptive state, and distributed tactile measurements. Demonstration data and SONIC policy updates are recorded at 50~Hz, while the low-level motor command loop runs at 500~Hz.

The main text describes the data scale as approximately 50 demonstrations per task. This is the nominal collection scale; after task-specific collection and cleaning, the retained datasets contain 52--63 episodes per task. Table~\ref{tab:dataset_stats} reports the exact retained statistics used for the final dataset collection.

\begin{table}[H]
\centering
\small
\caption{Retained demonstrations. Duration is computed from the 50~Hz dataset frames.}
\label{tab:dataset_stats}
\begin{tabular*}{\columnwidth}{@{\extracolsep{\fill}}lrrr@{}}
\toprule
Task & Episodes & Frames & Duration \\
\midrule
Table Sweeping & 56 & 72,301 & 24:06 \\
Back-Tap Walking & 63 & 54,405 & 18:08 \\
Human--Robot Hugging & 62 & 44,531 & 14:51 \\
Basket Loading & 56 & 69,004 & 23:00 \\
Composed Cleanup & 52 & 139,156 & 46:23 \\
\midrule
Total & 289 & 379,397 & 2:06:28 \\
\bottomrule
\end{tabular*}
\end{table}

The retained datasets are stored on a 50-Hz observation grid. The dataset metadata defines a training split containing all retained episodes, without a separate validation split; checkpoints are selected based on real-world rollout performance. We preserve episode boundaries when forming training windows so that tactile histories, future prediction targets, and action chunks never cross trajectory boundaries.

\subsection{Language Instructions}
\label{app:instructions}

Each task uses the following instruction, reproduced from the retained dataset metadata:

\noindent\textbf{Table Sweeping:} Use your forearm to sweep all objects across the blue divider from one side of the table to the other.\par\smallskip
\noindent\textbf{Back-Tap Walking:} Walk forward while you feel continuous contact or pushing on your back, and stop when the contact is released.\par\smallskip
\noindent\textbf{Human--Robot Hugging:} Walk up to the person, gently hug them, hold briefly, and release.\par\smallskip
\noindent\textbf{Basket Loading:} Hold the basket steady while the person loads objects into it.\par\smallskip
\Needspace{5\baselineskip}
\noindent\textbf{Composed Cleanup:} Sweep the garbage from the high table into the storage box on the adjacent lower table using your forearm. Then pick up the box by hugging it with both arms, carry it to the black trash bin, and throw the entire box into the bin.\par

\subsection{Multimodal Observations and Tactile Sensing}
\label{app:sensing}

The robot uses a forward-facing stereo RGB camera. The recorded dataset stores the left and right views separately at $640\times480$ and 50~fps. For the Isaac-GR00T implementation, each view is padded to a square, resized to 256 pixels on the short edge, cropped to 95\% of the spatial extent, and resized to $256\times256$; random cropping is used during training and center cropping during deployment.

The proprioceptive policy input has 46 dimensions: 29 body joint positions, 14 left/right hand-interface placeholder dimensions retained for compatibility, and 3 projected-gravity coordinates. The evaluated G1 does not use physical dexterous hands; the placeholder dimensions are part of the policy interface rather than additional actuated joints.

Whole-body contact is measured using custom piezoresistive textile electronic skin from JQ Industries. The system consists of one torso garment and two arm sleeves. The torso garment exposes 112 effective transmitted taxels and each arm sleeve exposes 256, giving 624 effective tactile channels in total. The manufacturer specifies 8-bit ADC values in $[0,255]$ and a nominal sensor sampling rate of 100~Hz for both wired and Bluetooth acquisition. In our dataset pipeline, the latest available tactile measurements are sampled onto the common 50~Hz observation grid.

We decompose the tactile channels into eight regions: six torso/shoulder patches and two arm sleeves. Before learned encoding, each stream is baseline-corrected and small residual values are suppressed by a fixed deadband. The four-frame causal tactile history corresponds to the current observation and the preceding 60 ms on the 50-Hz dataset grid. The four future predictive targets correspond to nominal offsets of 20, 40, 60, and 80 ms.

For the arm sleeves, the Isaac-GR00T implementation pools measurements around the sleeve circumference while preserving the axial contact profile. This reduces sensitivity to sleeve rotation while retaining where contact occurs along the arm. The $\pi_{0.5}$ implementation uses the same selected tactile regions and four-frame causal history but retains its backbone-specific tactile preprocessing and feature width.

\section{Training and Deployment Details}
\label{app:training}

\subsection{Backbone Adaptation and Tactile Modules}
\label{app:backbone}

For Isaac-GR00T, we keep the pretrained vision-language model frozen and fine-tune the action expert, state encoder, tactile pathway, and predictive modules. The tactile pathway first applies a region-specific MLP, maps the regional features into eight learned-query tactile tokens through 8-head cross-attention, and then models the four-frame history with a temporal Transformer. The temporally encoded tactile tokens are inserted into the action expert together with the state token and noisy action tokens. After the DiT trunk, the contextualized tactile-token outputs are mean-pooled to form the tactile-anchored multimodal context used by the three predictive heads.

In the Isaac-GR00T implementation, the regional MLP maps each selected tactile region through $d_r\!\rightarrow\!512\!\rightarrow\!1536$. The temporal module projects tactile tokens from 1536 to 512 dimensions, applies one 8-head Transformer layer with a 1024-dimensional feed-forward block, and projects the output back to the policy width. The post-DiT tactile context is 1024-dimensional. The tactile and proprioceptive predictors use $1024\!\rightarrow\!512\!\rightarrow\!4\times1536$ MLPs, while the visual predictor uses $1024\!\rightarrow\!512\!\rightarrow\!4\times2048$.

The $\pi_{0.5}$ adaptation follows the same overall design---eight tactile tokens, a four-frame temporal history, post-action-expert tactile readout, and four-step multimodal prediction---while using the native 1024-dimensional action-expert space. Its regional MLP is $d_r\!\rightarrow\!512\!\rightarrow\!1024$, and the temporal block operates directly at width 1024 with an 8-head attention layer and a $1024\!\rightarrow\!2048\!\rightarrow\!1024$ feed-forward block. The tactile and proprioceptive predictors output four 1024-dimensional targets; the visual predictor outputs four 2048-dimensional targets.

For both backbones, the tactile and proprioceptive target encoders are updated by exponential moving average with decay 0.99, while the visual target is extracted by the frozen image encoder. Predictive targets are used only during training. The predictive heads and target encoders are not required for deployed action generation.

\subsection{Optimization}
\label{app:optimization}

Isaac-GR00T models and ablations are trained on four NVIDIA H100 GPUs, while $\pi_{0.5}$ models are trained on eight NVIDIA H100 GPUs. All models are optimized for 50,000 steps. Table~\ref{tab:training_hparams} summarizes the principal optimization settings. Ablations follow the Isaac-GR00T settings unless the ablated component itself is changed.

\begin{table}[H]
\centering
\small
\caption{Training settings for the two pretrained VLA backbones.}
\label{tab:training_hparams}
\begin{tabular}{lcc}
\toprule
Setting & Isaac-GR00T & $\pi_{0.5}$ \\
\midrule
Optimizer & AdamW & AdamW \\
Peak learning rate & $1\times10^{-4}$ & $5\times10^{-5}$ \\
LR schedule & cosine & warmup + constant \\
Warmup & 5\% & 1,000 steps \\
Weight decay & $1\times10^{-5}$ & $1\times10^{-10}$ \\
Global batch size & 64 & 64 \\
Gradient clipping & 1.0 & 1.0 \\
Training steps & 50,000 & 50,000 \\
GPUs & 4$\times$H100 & 8$\times$H100 \\
Precision & bfloat16 & bfloat16 \\
Teacher EMA & 0.99 & 0.99 \\
Seed & 42 & 42 \\
\bottomrule
\end{tabular}
\end{table}

The multimodal predictive objective uses $\lambda_T=0.05$ and $\lambda_P=\lambda_V=0.005$, with $\beta=1$ for the SmoothL1 latent-magnitude term. All three predictors output $K=4$ future latent targets. Stop-gradient is applied to the target representations, so predictive losses update the online policy/tactile pathway and prediction heads but not the target branches directly.

\subsection{Deployment}
\label{app:deployment}

At deployment, the policy predicts a chunk of $H_A=40$ 64-dimensional SONIC motion tokens. Commands are consumed on the 50~Hz control grid, so a full chunk spans 0.8~s. The public deployment implementation triggers a new policy request approximately 0.4~s after receiving the previous result (nominal request rate 2.5~Hz); while a new prediction is pending, execution continues from the current chunk. When the new chunk arrives, latency compensation skips the prefix corresponding to elapsed inference time before the new chunk is activated. Thus, deployment does not assume a fixed number of executed tokens per prediction.

The default deployment configuration does not use action-chunk blending, tactile-input EMA smoothing, or a multi-chunk temporal ensemble. The predictive heads and EMA target encoders are removed from the deployed computation path. Selected motion tokens are passed through SONIC Protocol v4 to the pretrained whole-body controller, which combines them with robot feedback and produces joint-position targets for the 500~Hz low-level motor-control loop.

\section{Detailed Evaluation Protocol}
\label{app:evaluation}

\subsection{General Protocol}
\label{app:general_eval}

All reported metrics are obtained from real-robot rollouts. Main-table Isaac-GR00T configurations use 20 rollouts per task and method. Cross-policy Isaac-GR00T success rates reuse the corresponding main-table evaluation, while each $\pi_{0.5}$ success-rate configuration uses 10 rollouts. In the ablation study, Full Uni-VLaT reuses the 20-rollout result and each other ablation uses 10 rollouts. Instability, falling, emergency stopping, or human assistance beyond the prescribed task interaction counts as failure. Prescribed human interaction---such as touching the robot's back, loading the basket, or participating in a hug---is part of the task and is not counted as intervention.

Tactile sensors remain mounted and are recorded during all evaluations, including No Tactile rollouts. For No Tactile, these measurements are withheld from the policy input and are used only for post-hoc analysis.

\subsection{Task-Specific Setup and Success Criteria}
\label{app:task_eval}

The initial object placement and success criteria for each task are specified below. Objects are randomized during demonstration collection and evaluation where noted. A trial that exceeds the stated timeout counts as failure.

\noindent\textbf{Back-Tap Walking.} The robot starts at one side of a designated laboratory walking area. As it traverses the area, a person alternates contact and release on its back 1--4 times. Success requires stable whole-body motion, forward movement within 1~s of contact, and stopping within 1~s of release, all within a 30~s rollout.

\noindent\textbf{Table Sweeping.} Five objects---a lightweight gear-shaped object, a cup, a thin IC card, a thin cardboard card, and an irregular duck-shaped object---are arranged in two rows, with their positions randomized within each row. The nominal table height is 90~cm and varies by approximately $\pm15$~cm. Success requires all five objects to cross the blue divider without falling off the table. Large unsafe interactions with the table count as failure. The timeout is 60~s.\par

\noindent\textbf{Basket Loading.} A person hands the basket to the robot, then loads a randomized sequence of several lightweight plastic bowls, a heavy object, a plastic bottle, and a solid ceramic bowl. The number and order of objects vary across trials. The robot must actively secure the basket after handover, keep it stable as objects are loaded, avoid dropping it or its contents or displacing it substantially, and release it when the person takes it away. The timeout is 60~s.\par

\noindent\textbf{Human--Robot Hugging.} The robot approaches a participant from a standing start. Demonstrations involve one participant, while evaluation involves two; clothing varies across interactions. Success requires a smooth bilateral embrace without large oscillatory motions, followed by release of both arms when the participant disengages and a return to stable standing. The timeout is 30~s.\par

\noindent\textbf{Composed Cleanup.} Household rubbish is placed on a table with fixed geometry. The task consists of three sequential stages: (1) sweep all rubbish into the adjacent box without omission or dropping; (2) secure and lift the box with both arms; and (3) carry the box to the designated position beside the black waste bin. The rollout is limited to 180~s.\par

\subsection{Composed Cleanup Stage Score}
\label{app:stage_score}

For the cross-policy comparison, we additionally report partial progress on Composed Cleanup because binary end-to-end success does not distinguish where a long-horizon rollout fails. Let $c_{i,j}\in\{0,1\}$ denote completion of scored stage $j$ in rollout $i$. The three stages are sweeping all rubbish into the box, securing and lifting the box, and reaching the waste bin while carrying it. The normalized stage score is
\begin{equation}
S_{\mathrm{stage}}=\frac{100}{N}\sum_{i=1}^{N}\frac{1}{3}\sum_{j=1}^{3}c_{i,j}.
\end{equation}
Thus, each completed stage contributes $100/3$ points to an individual rollout. Depositing the box in the bin is required for end-to-end success but is not a fourth scored stage; SR remains a separate binary metric.

\subsection{Diffusion-Policy Safety Screening}
\label{app:dp_screening}

The from-scratch Diffusion Policy baseline is reported as N/A rather than 0\% SR because it did not enter physical rollout evaluation. Three checkpoints from different training stages were screened using the same deployment interface and failed SONIC's pre-execution action validity check before their predicted chunks were sent to the robot. We therefore do not assign those checkpoints a real-robot success rate.

\end{document}